\documentclass[letterpaper]{article} 
\usepackage{aaai2026}  

\nocopyright

\usepackage{times}  
\usepackage{helvet}  
\usepackage{courier}  
\usepackage[hyphens]{url}  
\usepackage{graphicx} 
\usepackage{natbib}  
\usepackage{caption} 
\usepackage{algorithm}
\usepackage{algorithmic}
\usepackage{multirow} 
\usepackage{multicol}
\usepackage{array}
\usepackage{booktabs} 

\usepackage{newfloat}
\usepackage{listings}
\DeclareCaptionStyle{ruled}{labelfont=normalfont,labelsep=colon,strut=off} 
\floatstyle{ruled}
\newfloat{listing}{tb}{lst}{}
\floatname{listing}{Listing}
\usepackage{booktabs} 
\usepackage{multirow} 
\usepackage{amsmath}
\title{AutoSCORE: Enhancing Automated Scoring with Multi-Agent Large Language Models via Structured Component Recognition}
\author{
    Yun Wang\textsuperscript{\rm 1}\equalcontrib,
    Zhaojun Ding\textsuperscript{\rm 1}\equalcontrib,
    Xuansheng Wu\textsuperscript{\rm 1},
    Siyue Sun\textsuperscript{\rm 2},
    Ninghao Liu\textsuperscript{\rm 1},
    Xiaoming Zhai\textsuperscript{\rm 3}
}
\affiliations{
    \textsuperscript{\rm 1}School of Computing, University of Georgia \\
    \textsuperscript{\rm 2}Khoury College of Computer Sciences, Northeastern University \\
    \textsuperscript{\rm 3}AI4STEM Education Center, University of Georgia \\

    \{yun.wang1, zhaojun.ding, xuansheng.wu, ninghao.liu, xiaoming.zhai\}@uga.edu, sun.siyue@northeastern.edu
}

\usepackage{bibentry}

\begin{document}

\maketitle

\begin{abstract}
Automated scoring plays a crucial role in education by reducing the reliance on human raters, offering scalable and immediate evaluation of student work. While large language models (LLMs) have shown strong potential in this task, their use as end-to-end raters faces challenges such as low accuracy, prompt sensitivity, limited interpretability, and rubric misalignment. These issues hinder the implementation of LLM-based automated scoring in assessment practice. To address the limitations, we propose AutoSCORE, a multi-agent LLM framework enhancing \textbf{auto}mated scoring via rubric-aligned \textbf{S}tructured \textbf{CO}mponent \textbf{RE}cognition. With two agents, AutoSCORE first extracts rubric-relevant components from student responses and encodes them into a structured representation (i.e., Scoring Rubric Component Extraction Agent), which is then used to assign final scores (i.e., Scoring Agent). This design ensures that model reasoning follows a human-like grading process, enhancing interpretability and robustness. We evaluate AutoSCORE on four benchmark datasets from the ASAP benchmark, using both proprietary and open-source LLMs (GPT-4o, LLaMA-3.1-8B, and LLaMA-3.1-70B). Across diverse tasks and rubrics, AutoSCORE consistently improves scoring accuracy, human-machine agreement (QWK, correlations), and error metrics (MAE, RMSE) compared to single-agent baselines, with particularly strong benefits on complex, multi-dimensional rubrics, and especially large relative gains on smaller LLMs. These results demonstrate that structured component recognition combined with multi-agent design offers a scalable, reliable, and interpretable solution for automated scoring.
\end{abstract}


\section{Introduction}

Grades are an integral component of educational systems, serving to quantify learning, measure achievement, customize instruction for learners, motivate engagement, and provide individualized feedback ~\cite{cain2022deficiencies}. However, assessing and assigning these grades to constructed response assessments often relies on manual scoring, which is time-consuming for educators~\cite{Aniche2021}, failing to provide timely feedback for large classes or online settings. Moreover, human scoring does not always achieve ideal objectivity or inter-rater consistency~\cite{hussein2019}. These limitations highlight the need for scoring approaches that not only improve efficiency and scalability, but also enhance consistency and fairness in assessment. Automated scoring offers a way forward by reducing grading time, lowering costs~\cite{figueras2025promises}, and providing more timely feedback~\cite{nordquist2007providing, dikli2014automated}. Furthermore, under appropriate design and with sufficient training data, automated scoring can yield higher accuracy and potentially reduce certain biases compared to human raters~\cite{vo2023human, guo2025artificial}, thereby supporting fairness in assessment. As such, it is increasingly being adopted as part of modern educational practice, with large-scale assessments such as the TOEFL and GRE incorporating automated scoring engines to evaluate written responses alongside human raters~\cite{zhai2023large}.

While automated scoring has seen widespread uptake, several challenges remain. Firstly, traditional deep learning–based scoring models require substantial amounts of labeled data to achieve high accuracy, which limits their scalability across diverse tasks and contexts~\cite{ridley2020promptagnosticessayscorer}. Moreover, they often operate as black-box systems, offering little transparency into the specific patterns and features used for scoring, which makes it difficult for educators to trust~\cite{misgna2024survey}. In addition, the feedback they provide is often too generic to help students identify and address their most important areas for improvement, and it occasionally flags correct usage as errors or overlooks genuine mistakes~\cite{zhai2021meta}. Addressing these challenges calls for more advanced approaches that can generalize across diverse tasks, offer transparent decision-making, and deliver real-time feedback. 

One promising direction comes from recent advances in large language models (LLMs), which offer strong language understanding and generative capabilities~\cite{minaee2025largelanguagemodelssurvey} that can be leveraged to advance automated scoring, particularly for open-ended responses and essays~\cite{mansour2024can, sessler2025can, latif2024fine}. These models can generalize across prompts, enabling scoring on diverse tasks without extensive retraining~\cite{wu2023matching,lee2024unleashinglargelanguagemodels}. Beyond this, LLMs can explain their scoring decisions in natural language, improving transparency and interpretability for educators~\cite{lee2024applying,chu2025rationaleessayscoresenhancing}. In addition, LLMs can generate more targeted, task-specific feedback and reduce misjudgments~\cite{Scarlatos_2024}, representing a notable improvement over traditional deep learning-based scoring systems. However, despite these advantages, the end-to-end fashion of current LLM-based scoring still leads to several limitations: (i) Even machine's assessments have the same scores with humans, their reasoning paths may differ and be unclear or untraceable, making audit and justification difficult~\cite{wu2025unveiling}; (ii) Without an explicit step that checks each rubric criterion one by one, it is unclear whether LLM raters apply consistent standards across similar responses, consistency remains unverified~\cite{shi2025judgingjudgessystematicstudy}; (iii) Using a prompt that concatenates the rubric, question, and student response into a single input may cause uneven coverage of rubric dimensions, with some criteria overlooked and others overemphasized~\cite{zhu2025focusdirectionsmakelanguage}; (iv) Sensitivity to prompt or formatting variations, reducing reliability~\cite{errica2025didiwrongquantifying}.

We argue that these limitations arise from the absence of an explicit step that first identifies the evidence or components relevant to each rubric criterion before scoring. Without such a step, LLMs are more likely to deviate from criterion-grounded reasoning and produce inconsistent or incomplete evaluations. As a well-established tool in educational practice, rubrics enhance transparency by clarifying criteria~\cite{brookhart2018} and facilitate self-regulated learning, self-assessment, and self-efficacy~\cite{panadero2019using, andrade2019critical, brookhart2015quality}. In performance assessments, rubrics can also improve human scoring reliability and inter-rater agreement ~\cite{jonsson2007use, reddy2010review}. Therefore, if LLMs are viewed as ``rater,'' a more reasonable approach is not to assign scores all at once, but rather to align with the rubric, just as trained raters do: first identify evidence components corresponding to rubric items, then make judgments and assign scores, and finally produce auditable justifications linked to evidence.

Based on the above insights, we propose AutoSCORE, a rubric-guided multi-agent LLM scoring framework that explicitly identifies rubric-relevant contents from student's essays during scoring. This separated scoring process allows LLM raters to complete a distinct sub-task in each inference, where we call an LLM specified for a particular sub-task as an agent. AutoSCORE comprises two core agents: (i) a Scoring Rubric Component Extraction Agent that identifies structured components in student responses according to the scoring rubric and produces rubric-aligned outputs, and (ii) a Scoring Agent that assigns final scores based on these representations, the original responses, and the rubric. Optional Verification and Feedback Agents can be incorporated to denoise and align extraction results and to generate interpretable feedback, further enhancing quality control and transparency. We summarize our contributions as follows:

\begin{itemize}
\item Proposes a rubric-aligned structured scoring paradigm: first, scoring rubric component identification, then item-by-item scoring, forming an auditable evidence-judgment chain.
\item Designs and implements a model-agnostic multi-agent framework that can leverage both proprietary models (such as GPT-4) and high-performance open-source models, facilitating localization and cost control.
\item Evaluation is conducted on multiple datasets and question types, demonstrating improved consistency and robustness compared to single-step/single-model baselines.
\end{itemize}

\section{Related Work}

Automated scoring broadly refers to the use of automated methods to evaluate constructed-response tasks in educational assessments~\cite{zhai2020applying}. In the context of text-based assessment, prior research on automated scoring covers two major strands: automated essay scoring (AES), which evaluates extended writing tasks such as essays~\cite{shermis2013handbook}, and short answer scoring (SAS), which focuses on brief, content-specific responses~\cite{mohler2011learning}. Earlier approaches in both strands predominantly relied on handcrafted features and statistical models, which achieved moderate agreement with human ratings but required prompt-specific engineering and offered limited transparency.

With the advent of deep learning and neural representation learning in NLP, automated scoring methods for both AES and SAS shifted from handcrafted feature engineering to end-to-end models that learn task-specific features directly from text~\cite{taghipour2016neural, dong2017attention}. Early neural approaches employed recurrent or convolutional architectures to encode responses and predict scores~\cite{riordan2017investigating, taghipour2016neural, dong2017attention}, while more recent work leverages Transformer-based encoders, particularly BERT~\cite{devlin2019bert,latif2023automatic}, and pre-trained language models~\cite{zhang2019co, zhu2022automatic,liu2023context,latif2024g}. These models generally achieve higher agreement with human ratings and require less prompt-specific tuning, but often operate as black boxes and demand large labeled data~\cite{mayfield2020should, dong2017attention}. 

\begin{figure*}[t]
\centering
\includegraphics[width=0.95\textwidth]{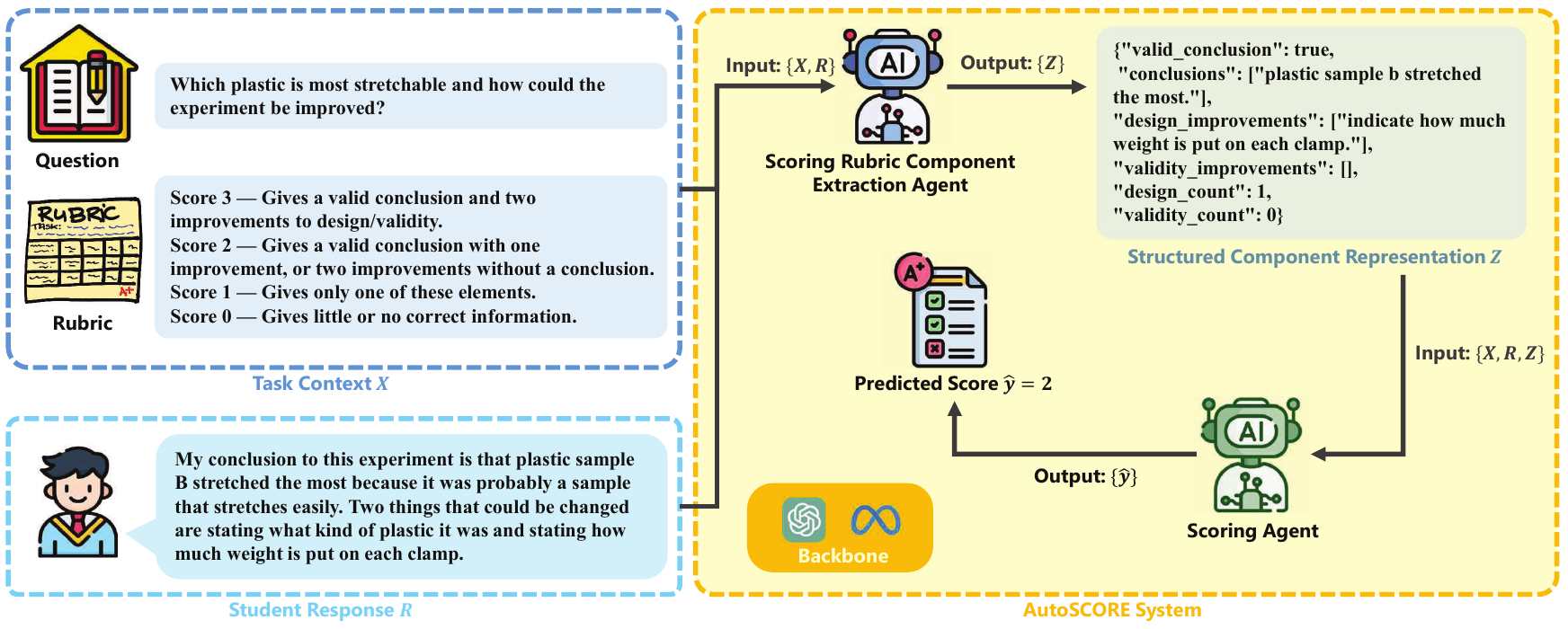} 
\caption{Overview of the proposed AutoSCORE multi-agent framework. The Scoring Rubric Component Extraction Agent identifies rubric-aligned components from the task context and student response, producing a structured representation $Z$. The Scoring Agent leveraged this representation together with the original inputs to assign the final score.}
\label{framework}
\end{figure*}

More recently, LLMs such as GPT-4~\cite{achiam2023gpt} and LLaMA~\cite{touvron2023llama} have demonstrated strong general-purpose language understanding and generation capabilities, enabling automated scoring in zero-shot or few-shot settings without extensive retraining~\cite{mansour2024can, sessler2025can,wu2023matching}. LLM-based scoring offers several advantages over traditional neural approaches: the ability to generate human-readable explanations of scores~\cite{chu2025rationaleessayscoresenhancing}, provide richer and more targeted feedback~\cite{Scarlatos_2024}, and generalize across prompts and tasks~\cite{lee2024unleashinglargelanguagemodels}. However, existing LLM-based approaches still face critical challenges: the reasoning path behind a score is often opaque or untraceable~\cite{wu2025unveiling}; without explicit criterion-level reasoning, it is unclear whether LLM raters apply consistent standards across similar responses~\cite{shi2025judgingjudgessystematicstudy}; criterion coverage can be unbalanced when scoring against rubrics~\cite{zhu2025focusdirectionsmakelanguage}; and outputs remain sensitive to prompt and formatting variations~\cite{errica2025didiwrongquantifying}.

Recent work has sought to address these limitations by incorporating rubrics into the scoring process, aiming to align evaluations more closely with human grading criteria and ensure more balanced coverage of rubric dimensions~\cite{wu2025unveiling, gunjal2025rubricsrewardsreinforcementlearning, eltanbouly2025trates, hashemi2024llm}. However, most existing rubric-guided approaches integrate rubric criteria only implicitly within the scoring prompt or model input, without explicitly identifying and structuring evidence for each criterion prior to scoring~\cite{wu2025unveiling, gunjal2025rubricsrewardsreinforcementlearning}. As a result, they may still exhibit inconsistencies in applying standards across responses and offer limited auditability of the reasoning process. Motivated by these limitations, our work adopts a rubric-aligned scoring paradigm that first extracts criterion-specific evidence before scoring, enabling greater consistency, interpretability, and traceability, and further leverages a multi-agent design to enhance reliability.

\section{Methodology}

We present AutoSCORE, a rubric-aligned, multi-agent LLM framework for automated scoring of constructed responses. It consists of two modules: Scoring Rubric Component Extraction Agent and Scoring Agent, which together ensure rubric alignment and interpretability. The architecture is shown in Figure~\ref{framework}.

\subsection{Problem Formulation}

We focus on the automated scoring task for assessing student-produced answers~\cite{williamson2012framework}, such as short answers and essays. Each entry is represented as a triplet $(X, R, y)$, where $X$ denotes the task context, including the assessment question, reference materials (e.g., data tables or figures), and the scoring rubric; $R$ is the student's free-form written response in the text format; and $y$ is the score assigned by human raters. The score $y$ comes from a finite ordinal set defined by the corresponding rubric (e.g. $\{0,1,2,3\}$). The objective is to design a system $f$ that, given $X$ and $R$, predicts a score
\[
\hat{y}=f_\theta(X, R),
\]
where $\hat{y}$ is expected to be as close as to the human assessment $y$. 
It is worth noting that, $f$ can be implemented with a single model or a framework with multiple LLMs. 

\subsection{Structured Component Representation}
To support rubric-grounded scoring, we propose an isolated process to check student responses according to the rubrics, rule-by-rule, before assigning the overall scores.
Specifically, AutoSCORE first extracts rubric-relevant \emph{components} from each student response $R$, and then encodes them into a structured format $Z$ to support the judgment of the overall score. 
Here, a component is defined as a piece of text from response $R$, reflecting a rule emphasized by the rubric $X$. 
Each component can be encoded into a different structured format, such as a Boolean flag, a count number, or the original text span itself.
For instance, consider a science investigation task scored with the following 3-point rubric $X$:
\begin{quote}
\textit{
Score 3: The response draws a valid conclusion and describes two ways to improve either the experimental design and/or the validity of the results.\\
Score 2: A valid conclusion with one improvement, or two improvements without a valid conclusion.\\
Score 1: Only one of these elements present.\\
Score 0: Little or no correct information from the investigation.
}
\end{quote}
This rubric consistently refers to three key components: (1) the presence of a valid conclusion, (2) improvements to the experimental design, and (3) improvements to the validity of the results. AutoSCORE first identifies them from a student response, and then organizes these components into a structured-format representation $Z$. Here, a \emph{representation} refers to a formatted record that explicitly restores rubric-relevant components extracted from the student responses, rather than a dense vector. 
Note that, if the format of $Z$ is loose or inconsistent, LLMs often fail to reuse this evidence to support their assessments reliably. To ensure consistency and human-readability, we adopt JSON as the encoding format for $Z$. For the rubric above, the components can be encoded in the JSON-style representation $Z$ as follows:
\begin{lstlisting}[numbers=none]
Z = {
  "valid_conclusion": true|false,
  "conclusions": [string, ...],
  "design_improvements": [string, ...],
  "validity_improvements": [string, ...],
  "design_count": integer,
  "validity_count": integer
}
\end{lstlisting}
In this example, we could observe that the three key components from the rubric are clearly defined as ``valid\_conclusion,'' ``design\_improvements,'' and ``validity\_improvements,'' respectively. Here, \texttt{true|false}, \texttt{string}, \texttt{integer} are data structures of the values that are used to store certain components for each keyword.

\subsection{Multi-Agent Framework}
To separate evidence extraction from score assignment and improve scoring accuracy and interpretability, AutoSCORE designs a multi-agent LLM framework, in which one agent identifies rubric-relevant components and encodes them into the structured representation $Z$ from the student response, and another agent determines the final score based on $Z$, the original response, and the rubric. This step-by-step design constrains the reasoning path of scoring through explicit checkpoints, making the decision process more transparent and enabling error isolation compared to single-pass scoring~\cite{trirat2024automl}.

\paragraph{Scoring Rubric Component Extraction Agent.} 
To explicitly capture the evidence required by the rubric, AutoSCORE employs a Scoring Rubric Component Extraction Agent. The motivation is that scoring directly from free-form responses often leads to missed or inconsistently treated criteria, lacking transparency. Instead, this agent identifies rubric-relevant components in the student response $R$, under the task context $X$, and encodes them into the structured representation $Z$. Formally, we have 
\[
Z = f_{\text{extract}}(X, R),
\]
which produces a structured representation $Z$, constrained by a prompt that enforces organize $Z$ as a valid JSON format, ensuring consistency and machine-readability. This explicit separation of evidence identification from scoring makes reasoning more transparent and provides rubric-aligned inputs for downstream scoring.

\paragraph{Scoring Agent.} 
The scoring agent assigns the final score $\hat{y}$ by leveraging the extracted representation $Z$, together with the task context $X$ and the original response $R$. 
By focusing on rubric-relevant evidence encoded in $Z$ and using $R$ for verification and error correction when inconsistencies arise, the agent ensures that predictions remain aligned with the rubric.
Formally, the scoring agent is defined as
\[
\hat{y} = f_{\text{scoring}}(Z, X, R),
\]
where $f_{\text{scoring}}$ predicts the final score $\hat{y}$. 
In particular, $f_{\text{scoring}}$ is asked to align with rubric guidelines, resolve ambiguities in favor of rubric definitions, and require strict integer-only JSON output. 
These requests further enhance interpretability of the scoring process and support targeted debugging. 

While our AutoSCORE framework can be extended with optional verification and feedback agents for additional quality control and interpretability, in this work, we focus on the two core agents described above to provide a clear evaluation of the core framework. We leave a detailed investigation of the optional modules for future work.

\subsection{Design Rationale: Reasoning as a Constrained Path}
The two-stage design of AutoSCORE can be viewed through a graph reasoning perspective. If we conceptualize the scoring process as a path from the task description to the final score, a human rater’s reasoning path passes through a sequence of intermediate checkpoints, i.e., identifying specific rubric-relevant components, before reaching at a decision. In this view, each checkpoint corresponds to a node in a reasoning graph, with multiple edges representing alternative reasoning steps. The Scoring Rubric Component Extraction Agent explicitly anchors these nodes by enforcing the extraction of rubric-grounded components into the structured representation $Z$. This constrains the LLM’s reasoning path to pass through human-aligned intermediate states.

Under the assumption that aligning these intermediate nodes increases the overlap between the model’s reasoning path and that of expert raters, this design improves scoring accuracy and interpretability to spurious correlations.

\section{Experiments and Analysis}

We evaluate our proposed AutoSCORE framework on four datasets and three LLMs, covering both short- and long-text scoring tasks while mitigating the randomness and model-specific bias of single-model evaluations.

\subsection{Experimental Setup}

\noindent\textbf{Datasets.}
Our primary experiments are conducted on the Automated Student Assessment Prize (ASAP) dataset, which was developed to evaluate whether computer systems can reliably score written responses for educational assessment. The dataset consists of two components: ASAP-SAS (Short Answer Scoring)~\cite{asap-sas} and ASAP-AES (Automated Essay Scoring)~\cite{asap-aes}. The ASAP-SAS dataset contains short student responses (fewer than 50 words) across multiple subjects, including \textbf{Science}, \textbf{Biology}, \textbf{English}, and other subjects,  with rubric score ranges of 0 to 2 or 0 to 3. It covers two response types: source-dependent and non-source-dependent. To ensure diversity in subject matter, rubric range, and response type, we select three subsets: \textbf{Science} (rubric 0–3, source-dependent), \textbf{Biology} (rubric 0–3, non-source-dependent), and \textbf{English} (rubric 0–2, source-dependent). All selected SAS responses are from grade 10 students.

To further evaluate the performance of the framework on long-text scoring, we conducted experiments on ASAP-AES \textbf{EssaySet}~\#1, where the average essay length is approximately 350 words. The dataset consists of responses from grade 8 students, with scores ranging from 1 to 6, and includes persuasive, narrative, and expository essays. 

In total, our evaluation covers 6,656 student responses: 4,871 from ASAP-SAS and 1,785 from ASAP-AES. 

\noindent\textbf{Model Selection.}
To evaluate the generality of our AutoSCORE framework, we experiment with both proprietary and open-source LLMs. The primary proprietary model is GPT-4o~\cite{hurst2024gpt}, chosen for in-depth evaluation because of its state-of-the-art performance across diverse reasoning and language understanding tasks, as well as its stable API access. 
To further examine the model-agnostic nature of our framework, we also include high-performing open-source LLMs that can be deployed locally. Specifically, we use LLaMA-3.1-8B-Instruct~\cite{dubey2024llama} and LLaMA-3.1-70B-Instruct, both of which have competitive performance in instruction-following and reasoning benchmarks. Evaluating across models of different origins and architectures allows us to verify that the benefits of the multi-agent framework are not limited to a single LLM.

\noindent\textbf{Baselines.}
We compare the proposed multi-agent scoring framework with a single-agent baseline for each selected LLM. In the baseline setting, the model is prompted once to directly generate a score for a given response according to the official scoring rubric. All other experimental conditions are identical between the two settings.

\noindent\textbf{Evaluation Metrics.}
The performance of our AutoSCORE framework was evaluated using six complementary metrics. Accuracy measures exact agreement with human raters, while Quadratic Weighted Kappa (QWK) captures weighted agreement by rewarding closer predictions and penalizing larger discrepancies. The Mean Absolute Error (MAE) and Root Mean Square Error (RMSE) quantify absolute and squared errors, and Pearson and Spearman correlations reflect linear and rank-based consistency with human scores.

\noindent\textbf{Implementation.}
All experiments were conducted on a workstation equipped with 3 × NVIDIA RTX A6000 GPUs and Intel Xeon Silver 4214R CPU @ 2.40GHz.

\subsection{Main Scoring Performance Comparison}

\begin{table*}[ht!]
\centering
\begin{tabular}{@{} l l cc cc cc @{}}
\toprule
\textbf{Datasets} & \textbf{Models} & \textbf{QWK}$\uparrow$ & \textbf{Accuracy}$\uparrow$ & \textbf{Pearson}$\uparrow$ & \textbf{Spearman}$\uparrow$ & \textbf{MAE}$\downarrow$ & \textbf{RMSE}$\downarrow$ \\
\midrule

\multirow{9}{*}{\shortstack[l]{Science subset \\ (3 Components)}}
 & GPT\!-4o & 0.701 & 0.588 & 0.707 & 0.696 & 0.451 & 0.733 \\
 & \quad + AutoSCORE & \textbf{0.717} & \textbf{0.632} & \textbf{0.728} & \textbf{0.718} & \textbf{0.418} & \textbf{0.726} \\
 & \quad$\Delta$ (\%) & +2.28\% & +7.48\% & +2.97\% & +3.16\% & -7.31\% & -0.96\% \\
 \cmidrule(lr){2-8}
 
 & LLaMA-3.1-8B-Instruct & 0.150 & 0.293 & 0.339 & 0.222 & 0.879 & 1.105 \\
 & \quad + AutoSCORE & \textbf{0.261} & \textbf{0.350} & \textbf{0.370} & \textbf{0.269} & \textbf{0.763} & \textbf{0.994} \\
  & \quad$\Delta$ (\%) & +74.00\% & +19.45\% & +9.15\% & +21.17\% & -13.20\% & -10.05\% \\
 \cmidrule(lr){2-8}
 
 & LLaMA-3.1-70B-Instruct & \textbf{0.533} & 0.426 & \textbf{0.617} & \textbf{0.542} & 0.663 & 0.925 \\
 & \quad + AutoSCORE & 0.504 & \textbf{0.462} & 0.543 & 0.431 & \textbf{0.588} & \textbf{0.837} \\
   & \quad$\Delta$ (\%) & -5.44\% & +8.45\% & -11.99\% & -20.48\% & -11.31\% & -9.51\% \\
  
\midrule

\multirow{9}{*}{\shortstack[l]{Biology subset \\ (1 Component)}}
 & GPT\!-4o & 0.681 & \textbf{0.819} & 0.695 & 0.687 & \textbf{0.188} & \textbf{0.450} \\
 & \quad + AutoSCORE & \textbf{0.743} & 0.806 & \textbf{0.750} & \textbf{0.695} & 0.198 & 0.453 \\
 & \quad$\Delta$ (\%) & +9.10\% & -1.59\% & +7.91\% & +1.16\% & +5.32\% & +0.67\% \\
 \cmidrule(lr){2-8}
 
 & LLaMA-3.1-8B-Instruct & \textbf{0.087} & 0.182 & \textbf{0.472} & \textbf{-0.230} & \textbf{0.826} & \textbf{0.918} \\
 & \quad + AutoSCORE & 0.063 & \textbf{0.184} & 0.422 & -0.255 & 0.827 & 0.921 \\
  & \quad$\Delta$ (\%) & -27.59\% & +1.10\% & -10.59\% & -10.87\% & +0.12\% & +0.33\% \\
 \cmidrule(lr){2-8}
 
 & LLaMA-3.1-70B-Instruct & \textbf{0.704} & \textbf{0.816} & \textbf{0.709} & 0.517 & \textbf{0.188} & \textbf{0.445} \\
 & \quad + AutoSCORE & 0.660 & 0.739 & 0.699 & \textbf{0.570} & 0.278 & 0.560 \\
   & \quad$\Delta$ (\%) & -6.25\% & -9.44\% & -1.41\% & +10.39\% & +47.87\% & +25.84\% \\
 
\midrule

\multirow{9}{*}{\shortstack[l]{English subset \\ (4 Components)}}
 & GPT\!-4o & 0.540 & 0.548 & 0.603 & 0.501 & 0.468 & 0.708 \\
 & \quad + AutoSCORE & \textbf{0.629} & \textbf{0.604} & \textbf{0.653} & \textbf{0.553} & \textbf{0.398} & \textbf{0.633} \\
  & \quad$\Delta$ (\%) & +16.48\% & +10.22\% & +8.29\% & +10.38\% & -14.96\% & -10.59\% \\
 \cmidrule(lr){2-8}
 
 & LLaMA-3.1-8B-Instruct & 0.354 & 0.430 & 0.482 & 0.408 & 0.628 & 0.863 \\
 & \quad + AutoSCORE & \textbf{0.447} & \textbf{0.492} & \textbf{0.505} & \textbf{0.449} & \textbf{0.547} & \textbf{0.790} \\
  & \quad$\Delta$ (\%) & +26.27\% & +14.42\% & +4.77\% & +10.05\% & -12.9\% & -8.46\% \\
 \cmidrule(lr){2-8}
 
 & LLaMA-3.1-70B-Instruct & 0.422 & 0.463 & 0.564 & 0.439 & 0.568 & 0.794 \\
 & \quad + AutoSCORE & \textbf{0.490} & \textbf{0.532} & \textbf{0.572} & \textbf{0.470} & \textbf{0.486} & \textbf{0.721} \\
   & \quad$\Delta$ (\%) & +16.11\% & +14.90\% & +1.42\% & +7.06\% & -14.44\% & -9.19\% \\

\midrule

\multirow{9}{*}{\shortstack[l]{Essay Set\\ (5 Components)}}
 & GPT\!-4o & 0.251 & \textbf{0.280} & 0.412 & 0.343 & 1.076 & 1.396 \\
 & \quad + AutoSCORE & \textbf{0.344} & 0.269 & \textbf{0.499} & \textbf{0.453} & \textbf{1.035} & \textbf{1.327} \\
  & \quad$\Delta$ (\%) & +37.05\% & -3.93\% & +21.12\% & +32.07\% & -3.81\% & -4.94\% \\
 \cmidrule(lr){2-8}
 
 & LLaMA-3.1-8B-Instruct & 0.135 & 0.128 & 0.361 & \textbf{0.301} & 1.364 & 1.596 \\
 & \quad + AutoSCORE & \textbf{0.193} & \textbf{0.165} & \textbf{0.436} & 0.273 & \textbf{1.222} & \textbf{1.451} \\
  & \quad$\Delta$ (\%) & +42.96\% & +28.91\% & +20.78\% & -9.30\% & -10.41\% & -9.09\% \\
 \cmidrule(lr){2-8}
 
 & LLaMA-3.1-70B-Instruct & 0.428 & 0.340 & 0.599 & 0.525 & 0.862 & 1.143 \\
 & \quad + AutoSCORE & \textbf{0.575} & \textbf{0.488} & \textbf{0.649} & \textbf{0.549} & \textbf{0.612} & \textbf{0.914} \\
   & \quad$\Delta$ (\%) & +34.35\% & +43.53\% & +8.35\% & +4.57\% & -29.00\% & -20.03\% \\
   
\bottomrule
\end{tabular}
\caption{Main scoring performance comparison across datasets, models, and evaluation metrics, highlighting the impact of AutoSCORE over baseline models. Bold numbers denote the best performance within each dataset–metric pair.}
\label{maintable}
\end{table*}

Table \ref{maintable} summarizes the main results across four datasets and three LLMs. Overall, AutoSCORE achieves consistent improvements in most dataset–metric combinations compared to single-agent baselines. For instance, on Science subset with GPT-4o, Accuracy rises from 0.588 to 0.632 (+7.5\%), while on English subset, QWK improves from 0.540 to 0.629 (+16.5\%). On the long-text task Essay Set, QWK increases from 0.251 to 0.344 (+37.1\%) with Accuracy also rising by 26.9\%. Although a few settings show marginal declines (e.g., QWK for the LLaMA-3.1-8B model on the Biology subset), the overall trend across Accuracy, QWK, MAE/RMSE, and correlations indicates that rubric-aligned multi-agent scoring provides consistent benefits, particularly for tasks requiring complex rubric coverage.

\noindent\textbf{Task-Level Analysis.} 
Performance improvements vary with the complexity of both the rubric and the underlying task. On Biology subset, where scoring essentially reduces to counting key elements, the baseline already aligns well with the rubric. Here AutoSCORE shows limited benefits: GPT-4o improves slightly (QWK +9.1\%), while the LLaMA-3.1-8B model even declines. This may be due to (i) a ceiling effect, where the task is too simple to benefit from extra structure, and (ii) error propagation, where noise in component extraction can outweigh potential benefits.
In contrast, the English subset and Essay Set involve more complex responses and multi-dimensional rubrics. These settings require accurate extraction of rubric-relevant components to support scoring, where AutoSCORE yields the largest and most consistent improvements. For example, on Essay Set, QWK increases by +37.1\%. These results confirm that the framework is particularly valuable for tasks with complex rubrics or longer responses, where accurate component recognition becomes critical for reliable scoring.

\noindent\textbf{Model-Level Analysis.}
We examine how model capacity affects the relative benefits of AutoSCORE. The relative gains are larger for smaller models. On Science subset, the LLaMA-3.1-8B model improves QWK from 0.150 to 0.261 (+74.0\%), whereas GPT-4o achieves a much smaller relative gain (+2.3\%) under the same task. Similar patterns are observed on English subset, where the relative gain is +26.3\% for the LLaMA-3.1-8B model compared to +16.5\% for GPT-4o, and on Essay Set (+43.0\% for LLaMA-3.1-8B model and +37.1\% for GPT-4o). This indicates that AutoSCORE is especially effective when the backbone model has limited capacity: by decoupling component recognition from final scoring, AutoSCORE compensates for weaker reasoning ability. Practically, this implies that institutions with limited computational resources can deploy smaller models and still achieve substantial accuracy gains.

\begin{table*}[t]

\begin{tabular}{>{\raggedleft\arraybackslash}p{3.5cm} p{13.5cm}} 
\toprule
\multicolumn{2}{l}{\textit{Human Score = 1},  \textit{AutoSCORE = 1},  \textit{Baseline = 0}}\\
\textbf{Assessment Question} & How does the author organize the article? Support your response with details from the article. \\
\textbf{Student Response} & The author organizes the article in parts. He starts with an introduction to pull the reader in and then quickly changes the tone to show that he is still taking the article seriously. \\
\textbf{Rubric Excerpt} & 
\textbf{1 pt (Partially Proficient):} Fulfills some requirements, but may be general or simplistic. \newline
\textbf{0 pt (Not Proficient):} Inaccurate, incomplete, or missing information. \\
\textbf{Selected Extracted Components} &
\textbf{Organization Method:} The author organizes the article in parts. \newline
\textbf{Supporting Details:} (i) He starts with an introduction to pull the reader in. (ii) Then quickly changes the tone to show seriousness. \\
\bottomrule
\end{tabular}
\caption{Case study example from English subset.}
\label{case-study}
\end{table*}

\subsection{Ablation and Robustness}
The single-agent scoring baseline can be viewed as an ablation that removes the Scoring Rubric Component Extraction Agent, collapsing the framework into direct end-to-end scoring. The consistent performance gap between AutoSCORE and this baseline across datasets quantifies the contribution of structured component recognition. Moreover, the improvements appear not only in correlation-based metrics (QWK, Pearson, Spearman) but also in error-based metrics (MAE, RMSE), suggesting robustness across evaluations.

\subsection{Validation of Component Recognition Reliability}
To assess the reliability of the component recognition agent, we conducted a double-annotation study on two ASAP-SAS subsets: Science subset (multi-component rubric) and Biology subset (key-element counting). For each dataset, we randomly sampled 20\% of responses (n=258 for Science subset and n=370 for Biology subset). Two trained annotators independently identified rubric-relevant components, with adjudication yielding a gold reference.

On Science subset, the agent achieved strong reliability on the binary valid conclusion label (accuracy 0.899, $F_1$ 0.918, Cohen’s $\kappa$ 0.788). For component counts, the agent reached MAE 0.295, RMSE 0.564, Pearson $r$ 0.688 for design\_count, and MAE 0.116, RMSE 0.374, Pearson $r$ 0.777 for validity\_count. The exact-match rates were 0.717 and 0.895, respectively, confirming moderate-to-high fidelity in recovering rubric-relevant counts. While minor deviations remained in exact matches, the low error and moderate-to-high correlation suggest that the agent provides dependable component counts for subsequent scoring.

On Biology subset, where the rubric reduces to enumerating key elements, predictions were even more stable. The agent achieved an exact-match rate of 0.859, with MAE 0.157, RMSE 0.441, and a Pearson correlation of 0.893 against human annotations. The higher correlation and lower error in this setting reflect the simpler rubric structure, confirming that the agent can reproduce key-element counts.

Taken together, these findings demonstrate that the Component Recognition Agent reliably extracts rubric-grounded components across tasks of varying complexity. This provides a solid foundation for the AutoSCORE framework: accurate intermediate representations can be trusted as inputs to the downstream scoring stage, ensuring that observed improvements in overall scoring performance are not undermined by unreliable component recognition.

\subsection{Averaged Inference Time and QWK Tradeoff}

To better analyze the changes in both efficiency and performance after applying our proposed AutoSCORE framework, we conducted experiments on the English subset as a representative case. Specifically, we evaluated the tradeoff between QWK, a widely used performance metric in automated scoring, and the averaged inference time per instance, which reflects the efficiency of LLM-based scoring systems. Three models were tested: GPT-4o, LLaMA-3.1-70B-Instruct, and LLaMA-3.1-8B-Instruct.

\begin{figure}[t]
\centering
\includegraphics[width=0.95\columnwidth]{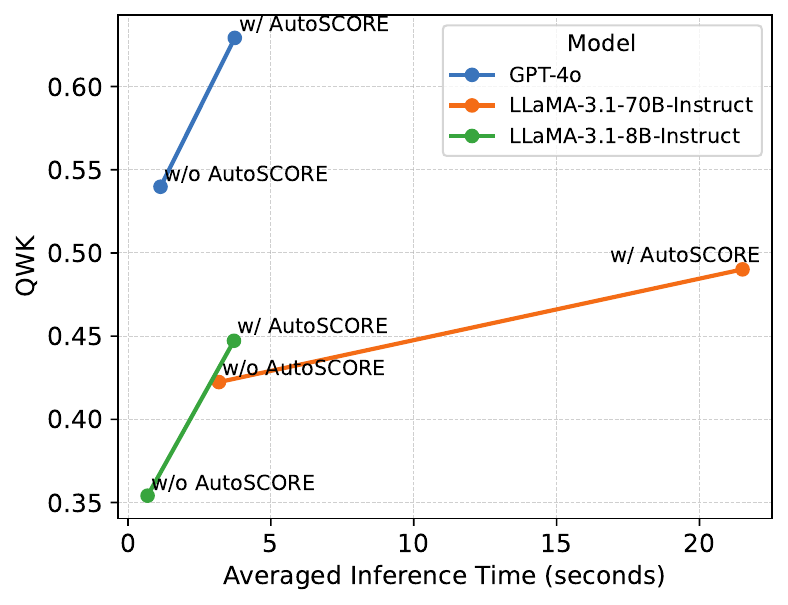} 
\caption{QWK compared with averaged inference time across different LLMs with and without AutoSCORE.}
\label{inferencetime}
\end{figure}

As shown in Figure \ref{inferencetime}, the results consistently demonstrate that incorporating AutoSCORE leads to higher QWK values at the cost of increased inference time across all three models. This highlights a clear efficiency–performance tradeoff: while AutoSCORE requires more time per inference, it yields more reliable scoring outcomes. Importantly, AutoSCORE provides a practical pathway to improving performance without necessitating access to substantially larger open-source models (which demand more computational resources) or more expensive proprietary models.


\subsection{Case Study}
To illustrate why our proposed AutoSCORE framework achieves more reliable scoring than directly applying an LLM to assign scores, we conduct a case study on the English subset. In this dataset, students were asked to read an article about space junk and respond to the assessment question about ``How does the author organize the article?'' 




As shown in Table \ref{case-study}, the rubric differentiates between responses that merely provide vague or incomplete statements and those that articulate how the article is organized with at least some supporting detail. In this example, the student identified both the organizational structure and supporting evidence. AutoSCORE explicitly extracted these rubric-relevant components, guiding the LLM to follow a reasoning path aligned with human raters and yielding the correct score of 1. In contrast, when GPT-4o scored the response directly, it overlooked key details and produced a score of 0. This comparison highlights how AutoSCORE constrains the model's reasoning through rubric-grounded intermediate states, thereby improving scoring accuracy.

\section{Conclusion and Discussion}

In this work, we introduced AutoSCORE, a rubric-aligned multi-agent LLM framework for automated scoring. By explicitly separating component extraction from scoring, AutoSCORE constrains the reasoning trajectory of LLMs to pass through human-aligned intermediate states, thereby addressing key limitations of end-to-end scoring such as rubric misalignment and lack of interpretability. Our experiments on four benchmark datasets (three ASAP-SAS subsets and one ASAP-AES set) and three LLMs demonstrate that AutoSCORE consistently outperforms single-agent baselines across Accuracy, QWK, correlation metrics, and error reduction. Notably, the framework shows the largest gains on tasks with multi-dimensional rubrics and long-form responses, confirming that explicit component recognition is crucial for reliable scoring in complex educational assessments. We also observe that AutoSCORE yields especially large relative improvements on smaller LLMs, highlighting its practicality for cost-sensitive educational applications.


While AutoSCORE demonstrates consistent improvements across datasets and models, several limitations remain. First, our experiments primarily focus on non-reasoning LLMs, which may differ from reasoning-oriented models (e.g., ChatGPT-o1) and could align differently with rubric-guided scoring, and evaluating AutoSCORE on such models warrants further investigation. Second, although our experiments provide theoretical evidence of effectiveness, AutoSCORE has not yet been deployed in a real classroom or large-scale student assessment settings. Future validation in authentic classrooms will be crucial to understanding AutoSCORE’s potential impact on student learning, teacher workload, and assessment reliability. Moreover, our work is limited to text-based assessments, whereas modern education increasingly involve multimodal inputs like images and audio, and extending AutoSCORE to handle such modalities remains an important direction for future research.



\end{document}